\documentclass{article} 
\usepackage{iclr2027_conference,times}

\usepackage{amsmath,amsfonts,bm}

\def\eqref#1{equation~\ref{#1}}

\def\1{\bm{1}}

\DeclareMathAlphabet{\mathsfit}{\encodingdefault}{\sfdefault}{m}{sl}
\SetMathAlphabet{\mathsfit}{bold}{\encodingdefault}{\sfdefault}{bx}{n}

\title{DiffMath: Symbol- and Graph-Aware \\ Latent Diffusion Transformer for Handwritten Mathematical Expression Generation}

\author{\textbf{Wei Pan}\textsuperscript{1}\quad
\textbf{Xuhan Zheng}\textsuperscript{1}\quad
\textbf{Huiguo He}\textsuperscript{1}\quad
\textbf{Yilin Shi}\textsuperscript{1}\quad
\textbf{Hiuyi Cheng}\textsuperscript{1}\\
\textbf{Dezhi Peng}\textsuperscript{2}\quad
\textbf{Minghui Liao}\textsuperscript{2}\quad
\textbf{Lianwen Jin}\textsuperscript{1}\thanks{Corresponding author. Code will be available at \href{https://github.com/awei669/DiffMath}{https://github.com/awei669/DiffMath}.}\\
\textsuperscript{1}South China University of Technology \quad
\textsuperscript{2}Huawei Technologies Co., Ltd.\\
\texttt{eewpan@mail.scut.edu.cn, eelwjin@scut.edu.cn}
}

\usepackage{subcaption}
\usepackage{graphicx}
\usepackage{hyperref}
\usepackage{url}
\usepackage{algorithmicx}
\usepackage{cleveref}
\usepackage{wrapfig}
\usepackage{booktabs}   
\usepackage{array}      
\usepackage{pifont}     

\newcommand{\cmark}{\checkmark} 
\newcommand{\xmark}{\ding{55}}  
\usepackage{graphicx}
\usepackage{newfloat}
\usepackage{multirow}  
\usepackage{array}     
\usepackage{pifont}
\usepackage{caption}
\usepackage{adjustbox}
\usepackage{float}
\floatstyle{ruled}
\newfloat{Algorithm}{t}{lop}
\floatname{Algorithm}{Algorithm}
\usepackage{enumitem}
\usepackage{titletoc}
\usepackage{tabularx}
\usepackage{amsmath,amssymb,amsfonts}%
\usepackage{cleveref}
\newcolumntype{Y}{>{\raggedright\arraybackslash}X}
\usepackage{multirow}
\usepackage{makecell}
\usepackage{xcolor}
\usepackage{algpseudocode}

\iclrfinalcopy 
\begin{document}

\maketitle

\begin{abstract}

Handwritten Mathematical Expression Generation (HMEG) remains challenging because it requires accurate modeling of complex hierarchical structures and two-dimensional layouts. Existing methods typically rely on explicit spatial supervision to model expression layouts, resulting in high annotation costs and limited scalability. To address this, we propose \textbf{DiffMath}, a structure-aware latent diffusion framework that leverages structural priors derived from LaTeX, thereby eliminating the need for positional supervision.
First, we design a \textbf{Rel}ational \textbf{A}bstract \textbf{S}yntax \textbf{T}ree (\textbf{RelAST}), a generation-oriented structural representation that distills MathML trees into compact triplet sequences encoding symbol identities, spatial relations, and nesting depths. Building upon RelAST, we introduce \textbf{MathVAE}, which learns a structure-preserving latent space through symbol-aware and relation-aware perceptual regularization, jointly encoding symbol semantics and structural topology. Finally, \textbf{MathDiT} performs conditional denoising in the learned latent space conditioned on RelAST, further guided by a global symbol-count prior via Adaptive Layer Normalization (AdaLN) to improve structural coherence.
Experiments demonstrate that DiffMath generates structurally consistent handwritten mathematical expressions, consistently outperforms existing HMEG methods, and effectively improves downstream OCR performance through synthetic data augmentation.

\end{abstract}

\section{Introduction}\label{intro}
\begin{figure}[t]
    \centering
    \includegraphics[width=0.8\linewidth]{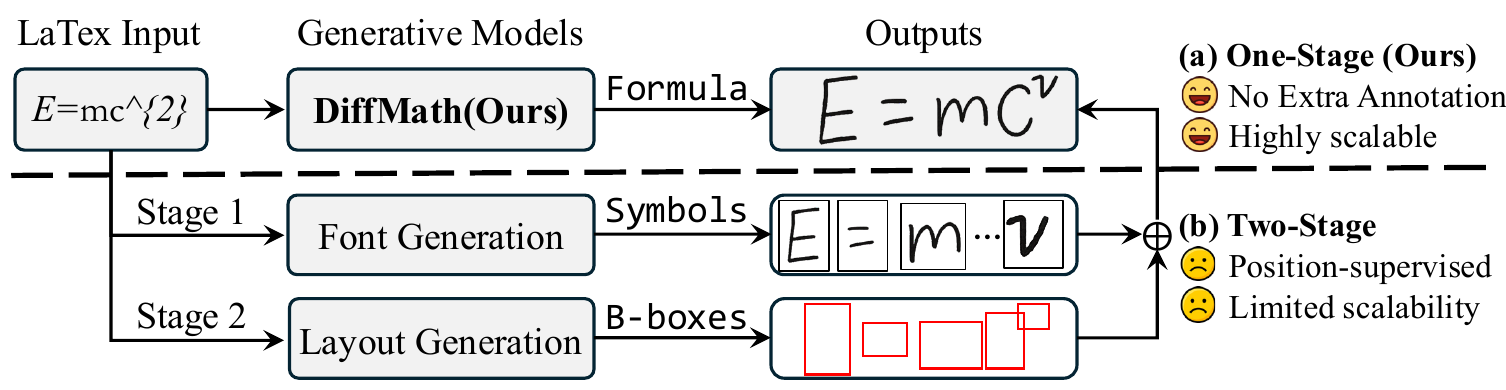}
    \vspace{-3mm}
    \caption{Unlike decoupled two-stage approaches that require explicit position-level supervision, \textbf{DiffMath (Ours)} adopts a streamlined end-to-end framework to directly map LaTeX to formula, reducing data dependency while improving global structural consistency.}
    \
    \label{fig:task_overview}
\end{figure}

Mathematical expressions are ubiquitous in education, scientific research, and are commonly represented using LaTeX. Unlike plain text, mathematical expressions exhibit rich hierarchical structures and spatial relationships, which can be compactly encoded by LaTeX through specialized syntax. Handwritten Mathematical Expression Generation (HMEG) aims to synthesize realistic handwritten trajectories of mathematical expressions from LaTeX strings, as illustrated in Fig.~\ref{fig:task_overview}(a). HMEG has broad applications in educational content creation and data augmentation for Handwritten Mathematical Expression Recognition (HMER). 
Compared with conventional augmentation methods~\cite{guan2024posformer} that mainly perturb existing handwritten samples, HMEG can synthesize diverse handwriting for previously unseen mathematical expressions, thereby expanding both content and structural diversity and providing more robust gains for downstream recognition.

\begin{table}[t]
    \centering
    \scriptsize
    \caption{Comparison of different LaTeX processing strategies for generative modeling in terms of syntax support, position modeling, and generation quality. Our method explicitly models positional information while supporting arbitrary LaTeX syntax and achieving better generation quality. ExpRate denotes expression accuracy, while FID measures distribution distance.}
    \vspace{-3mm}
    \label{tab:latex_prasing}
    \setlength{\tabcolsep}{5pt}
    \begin{tabular}{@{} l c cc cc @{}}
    \toprule
    \multirow{2.5}{*}{\textbf{Content Input}}
    & \multirow{2.5}{*}{\textbf{Position Modeling}}
    & \multicolumn{2}{c}{\textbf{Syntax Support}}
    & \multicolumn{2}{c}{\textbf{Generation Quality}} \\
    \cmidrule(lr){3-4} \cmidrule(lr){5-6}
    &
    & \textbf{Coverage}
    & \textbf{Extensible}
    & \textbf{ExpRate $\uparrow$}
    & \textbf{FID $\downarrow$} \\
    \midrule
    Raw LaTeX
    & None
    & Full
    & \checkmark
    & 56.27
    & 9.62 \\
    
    Syntax Tree~\citep{TAMER}
    & Explicit
    & Limited
    & \xmark
    & --
    & -- \\
    
    MathML~\citep{latexml}
    & Implicit
    & Full
    & \checkmark
    & 62.27
    & 7.57 \\
    \midrule
    \textbf{RelAST (Ours)}
    & Explicit
    & Full
    & \checkmark
    & \textbf{70.70}
    & \textbf{5.43} \\
    \bottomrule
    \end{tabular}
\end{table}

As a representative structured generation task, HMEG requires the generated expressions to be simultaneously content-correct, structurally accurate, and visually realistic. Unlike handwritten text generation, mathematical expressions exhibit complex two-dimensional layouts and long-range structural dependencies, where even slight structural errors may alter their semantic meaning. Some methods~\citep{Chen_2024_CVPR, gan2025stylized} therefore adopt a two-stage pipeline (Fig.~\ref{fig:task_overview}(b)) that separately generates symbols and predicts their spatial layouts. However, such decoupled modeling requires fine-grained layout annotations and often produces inconsistencies between symbol content and structural composition.

To overcome these limitations, we propose \textbf{DiffMath}, an end-to-end latent diffusion framework that directly learns HMEG from paired LaTeX and trajectory data without requiring additional layout annotations. Our key insight is that the performance of end-to-end HMEG critically depends on the structural representation used as the generation prior. As shown in Tab.\ref{tab:latex_prasing}, existing representations are not well suited for generation: raw LaTeX treats mathematical expressions as linear text, syntax-tree representations\citep{pmlr-v119-zhang20g} have limited syntax coverage, and standard MathML introduces redundant structural nodes that reduce information density. Motivated by these observations, we introduce Relational Abstract Syntax Tree (\textbf{RelAST}), a generation-oriented structural representation that explicitly encodes symbol identities, spatial relations, and structural hierarchy, providing compact and informative conditions for HMEG.

Building upon RelAST, DiffMath first employs \textbf{MathVAE}, a Variational Autoencoder that learns a structure-aware latent space for handwritten mathematical expressions. Inspired by DiffInk~\citep{pan2026diffink}, MathVAE extends reconstruction-based latent learning with symbol-aware and relation-aware regularization, encouraging the latent space to jointly preserve handwritten appearance, symbol semantics, and structural dependencies. The resulting structure-aware latent space provides an effective foundation for subsequent diffusion-based generation.

Building upon the learned structure-aware latent space, DiffMath further employs \textbf{MathDiT}, a conditional latent diffusion Transformer for HMEG. MathDiT performs denoising in the latent space while being conditioned on the explicit symbol, relation, and depth representations provided by RelAST. To further improve global structural consistency, we incorporate a symbol-count prior~\citep{can}, which is jointly embedded with the diffusion timestep through Adaptive Layer Normalization (AdaLN)~\citep{DiT}. By combining local structural conditions with global expression-level priors, MathDiT generates handwritten mathematical expressions with improved structural accuracy and visual fidelity.

Our contributions are summarized as follows:
1) We propose \textbf{DiffMath}, an end-to-end latent diffusion framework for HMEG that leverages structural priors derived from LaTeX, eliminating the need for dense positional supervision.
2) We design \textbf{RelAST}, a generation-oriented structural representation that transforms MathML trees into compact symbol--relation--depth triplets, providing expressive and efficient structural conditions for HMEG.
3) We introduce a structure-aware latent diffusion framework that learns structure-preserving latent representations through symbol-aware and relation-aware regularization, and further enhances generation with both local structural conditions and a global symbol-count prior.
4) Extensive experiments demonstrate that DiffMath achieves state-of-the-art performance on the MathWriting benchmark and that the generated samples consistently improve downstream HMER through synthetic data augmentation.
\vspace{-2mm}
\section{Related Works}
\label{sec:related_work}
\vspace{-2mm}
\textbf{Handwritten Text Generation} has progressed from RNN-based sequential modeling~\citep{graves2013generating} to Transformer- and diffusion-based frameworks~\citep{vatr,sdt,HWT,Pippi_2025_CVPR,DiffusionPen,OneDM,DiffBrush}. Recent methods~\citep{OLHWG,pan2026diffink} further improve generation quality through explicit layout modeling or structured latent diffusion. However, these approaches are designed for 1D linear text and are not well suited to mathematical expressions, which require modeling complex 2D topological structures.

\textbf{Handwritten Mathematical Expression Generation.}
Early methods~\citep{formula_gan} formulated HMEG as image translation, while subsequent approaches~\citep{Chen_2024_CVPR,gan2025stylized,wang2025sfrd} improved structural accuracy through explicit spatial modeling or style transfer. However, these methods rely on costly dense supervision, such as character-level bounding boxes or style labels. In contrast, DiffMath exploits structural priors derived from LaTeX parsing to model complex 2D topologies without positional annotations.

\textbf{Handwritten Mathematical Expression Recognition.}
Most HMER systems adopt encoder-decoder frameworks for handwritten mathematical expression recognition. Representative methods include BTTR~\citep{bttr}, CoMER~\citep{zhao2022comer}, CAN~\citep{can}, SAN~\citep{san}, TAMER~\citep{TAMER}, PosFormer~\citep{guan2024posformer}, and Uni-MuMER~\citep{li2025unimumer}, which progressively incorporate counting, syntax-aware decoding, hierarchical modeling, and vision-language learning. However, these tree representations retain redundant non-terminal container nodes that are unnecessary for generation.

\textbf{Text to Image Generation.}
Diffusion Transformers (DiTs)\citep{DiT} have become the dominant paradigm for high-fidelity generation and underpin many recent text-to-image systems\citep{podell2024sdxl,wu2025qwenimagetechnicalreport,team2025zimage,labs2025flux1kontextflowmatching,flux-2-2025}. Inspired by their success, we adopt a latent diffusion Transformer for HMEG, conditioned on a structured textual representation to better preserve symbolic correspondence and hierarchical mathematical layouts.

\vspace{-2mm}
\section{Method} 
\label{sec:method}

\vspace{-2mm}
\subsection{Preliminary}

\textbf{Data Format.}
A handwritten mathematical expression consists of a LaTeX expression and a trajectory sequence $X \in \mathbb{R}^{N \times 4}$. The LaTeX expression is a linear string encoding the symbolic structure of the formula. Each trajectory point is represented as $(x, y, pen)$, where $(x, y)$ denotes the spatial coordinates and $pen$ is a one-hot vector indicating the pen state (\textit{Pen Down} or \textit{Pen Up}).

\textbf{HMEG Task Overview.}
We formalize the task as follows: given a source LaTeX sequence, the goal is to generate an online handwriting trajectory $X$ that accurately renders the formula. The objective is to produce a sequence of points that not only captures the semantic essence of the mathematical expression but also adheres to the complex spatial-structural constraints inherent in its LaTeX representation.

\vspace{-2mm}
\subsection{Toward a Generation-Oriented Representation } 
\label{tree_parsing}
\textbf{Motivation.}
LaTeX linearizes mathematical logic into a structured 1D sequence, but often hides the 2D spatial relations among symbols, making it less effective when directly used as a generation condition. As summarized in Tab.~\ref{tab:latex_prasing}, raw LaTeX encodes layouts implicitly and leads to suboptimal generation quality. In HMER~\citep{li2025unimumer, TAMER, san, pmlr-v119-zhang20g}, syntax-tree parsing improves structural accuracy by explicitly modeling expression layouts. This observation suggests that structural parsing can also serve as a useful prior for generative modeling, where faithful layout synthesis is crucial. However, existing LaTeX processing strategies are not directly suited for generation: rule-based syntax trees have limited coverage, and MathML trees introduce redundant hierarchy without directly exposing spatial relations. These limitations motivate us to design RelAST, a compact generation-oriented representation that retains broad syntax coverage while explicitly encoding symbol identities, 2D relations, and hierarchical layouts.

\begin{figure}[t]
    \centering

    \begin{minipage}[t]{0.55\linewidth}
        \vspace{0pt}
        \centering
        \includegraphics[width=\linewidth]{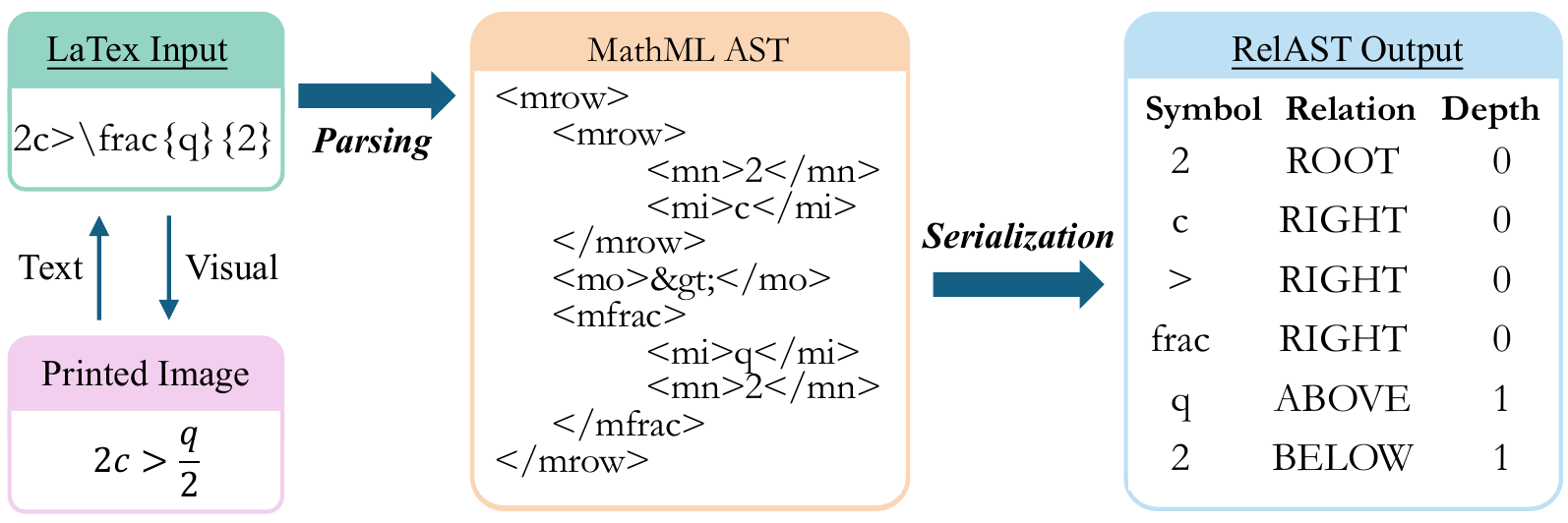}
        \vspace{-5mm}
        \caption{
        \textbf{RelAST construction.}
        LaTeX is converted into a standardized MathML tree and serialized
        via depth-first traversal into triplets $[S,R,D]$, representing
        symbol content ($S$), spatial relations ($R$), and hierarchical
        depth ($D$).
        }
        \label{fig:latex_parsing}
    \end{minipage}
    \hfill
    \begin{minipage}[t]{0.4\linewidth}
        \vspace{0pt}
        \centering
        \scriptsize
        \setlength{\tabcolsep}{2pt}
         \caption{
        Examples of different token types used in RelAST.
        }
        \renewcommand{\arraystretch}{1}

        \begin{tabularx}{\linewidth}{@{}lX@{}}
            \toprule
            \textbf{Token Type} & \textbf{Examples} \\
            \midrule

            Structural relations
            & \texttt{<RIGHT>, <SUP>, <SUB>, <ABOVE>, <BELOW>, <INSIDE>} \\

            Layout tokens
            & \texttt{frac, sqrt, root, table, <ROW>, <CELL>} \\

            Numbers
            & $0,1,2,3,4,5,\dots$ \\

            Latin variables
            & $x,y,z,a,b,c,A,B,C,\dots$ \\

            Greek symbols
            & $\alpha,\beta,\gamma,\theta,\lambda,\pi,\dots$ \\

            Arithmetic operators
            & $+,-,=,\times,\div,\cdot,\dots$ \\

            Calculus operators
            & $\int,\sum,\prod,\partial,\dots$ \\

            Delimiters
            & $(\ ),[\ ],\{\ \},|\,|,\dots$ \\

            \bottomrule
        \end{tabularx}

        \label{tab:token_types}

    \end{minipage}
\end{figure}

\textbf{LaTeX to RelAST.}
As illustrated in Fig.~\ref{fig:latex_parsing}, we first normalize LaTeX into a standardized MathML representation using LaTeXML~\citep{latexml} (\textbf{Step~1}), avoiding the limited coverage and inconsistent formats of rule-based parsers~\citep{li2025unimumer,TAMER,san,pmlr-v119-zhang20g}. We then remove non-terminal container nodes and perform depth-first pre-order traversal to obtain relational triplets $[S,R,D]$ (\textbf{Step~2}), where $S$, $R$, and $D$ denote the visible symbol, spatial relation, and hierarchical depth, respectively. The resulting representation explicitly preserves both 2D layouts and hierarchical structures, enabling the model to distinguish ambiguous expressions such as $a^{bc}$ and $a^b c$. The first relation is initialized as $R_{\mathrm{root}}$.
Representative examples of symbol and relation tokens are listed in Tab.~\ref{tab:token_types}.

\subsection{Overview of the DiffMath}
\begin{figure}[t]
    \centering
    \includegraphics[width=1\linewidth]{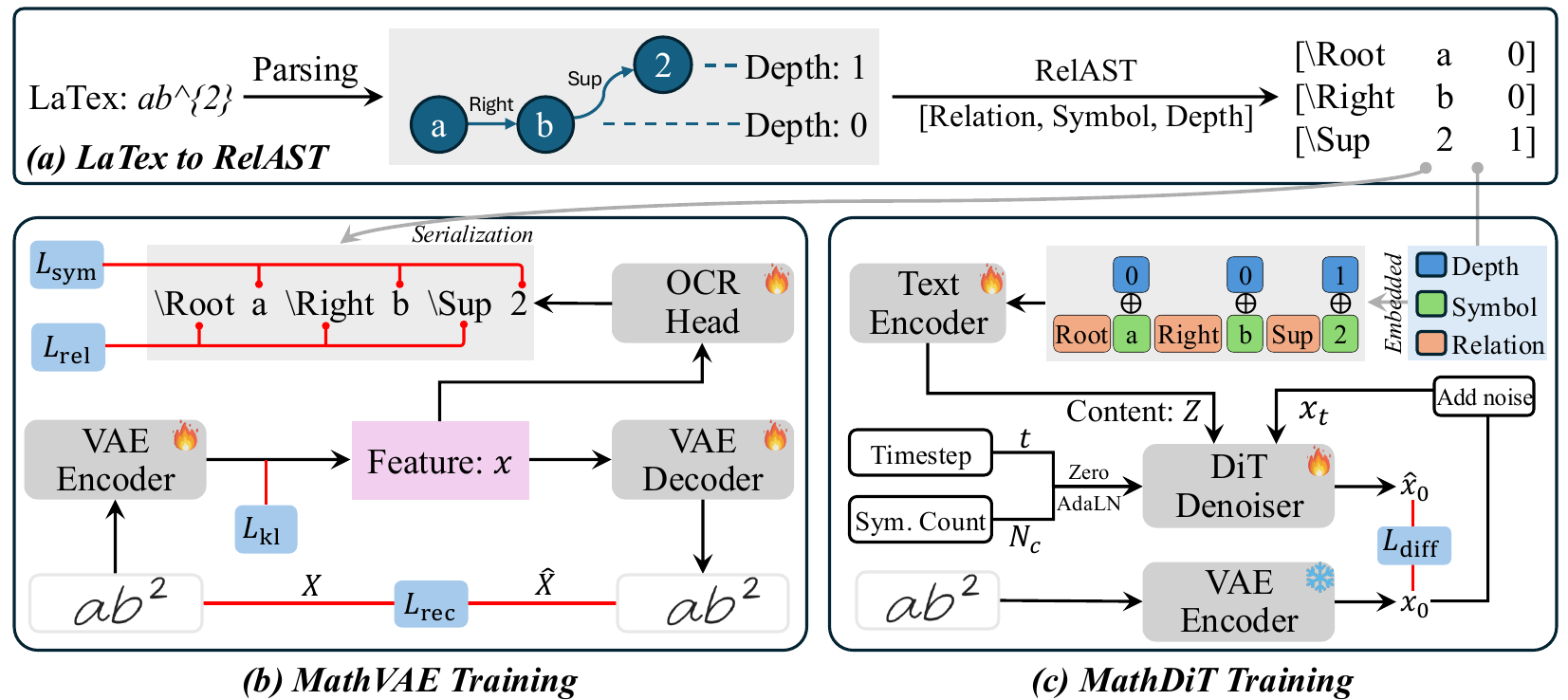}
    \caption{\textbf{Overview of the DiffMath Framework.}
    (a) LaTeX is first parsed into an explicit structured representation, termed RelAST.
    (b) MathVAE compresses raw trajectories into a latent spaceand introduces symbol- and relation-level perceptual losses ($\mathcal{L}_{\text{sym}}, \mathcal{L}_{\text{rel}}$) to enrich its semantic and structural representations.
    (c) MathDiT reconstructs the clean latent $\hat{x}_0$ from noise, conditioned on structural tokens and global counts for high-fidelity formula generation.}
    \label{fig_model_vae_dit}
\end{figure}

As shown in Fig.~\ref{fig_model_vae_dit}(a), input LaTeX is first parsed into RelAST, which encodes symbols, spatial relations, and nesting depths as structural guidance. Based on this representation, DiffMath consists of two main components: (1) MathVAE (Fig.~\ref{fig_model_vae_dit}(b)), a pre-trained sequential VAE for learning a structure-aware latent space; and (2) MathDiT (Fig.~\ref{fig_model_vae_dit}(c)), a conditional latent diffusion Transformer for structure-guided denoising.
MathVAE compresses raw online trajectories $X \in \mathbb{R}^{N \times 4}$ into latent representations $x \in \mathbb{R}^{l \times d}$ that preserve both symbol-level geometry and inter-symbol topology. In this latent space, MathDiT performs conditional denoising from $x_t$, which is obtained by perturbing $x$ during training or sampled from Gaussian noise during inference. Conditioned on the structural tokens $Z$ and the global symbol count $N_c$, MathDiT predicts the clean latent representation $\hat{x}_0$, which is decoded by MathVAE into the final handwritten mathematical expression.  

\subsection{Symbol- and Graph-Aware VAE.}
\label{sec:vae}
As shown in Fig.~\ref{fig_model_vae_dit}(b), MathVAE adopts an encoder--decoder architecture to map raw handwriting trajectories into a compact latent space. Rather than learning a purely reconstruction-oriented representation, MathVAE is designed to produce semantically informative latent features that facilitate downstream conditional diffusion. Motivated by recent studies on representation learning for generation and transfer~\citep{caron2021emerging,Oriane2026dinov,Yao_2025_CVPR,pan2026diffink,sclocchi2025probing}, we incorporate symbol- and relation-aware regularization, enabling the latent space to jointly encode handwritten appearance, symbol semantics, and spatial structure.

\vspace{1mm} 
\textbf{Sequence Encoder.}
To extract informative features from complex handwritten mathematical expressions, we employ a 1D convolutional encoder to transform the input handwriting trajectory sequence $X \in \mathbb{R}^{N \times 4}$ into a compact latent representation $x \in \mathbb{R}^{l \times d}$. This representation is then fed into two complementary branches: (1) a trajectory decoder for reconstructing low-level handwriting trajectories at the point level, and (2) a Transformer-based HMER module for capturing high-level semantic information, including symbolic content and structural relationships.

\vspace{1mm} 
\textbf{Sequence Decoder.}
The decoder first applies a 1D convolutional stack that mirrors the encoder, up-sampling the latent feature $x$ into a feature map $O_t \in \mathbb{R}^{N \times (6p + 2)}$. At each timestep, $O_t$ is divided into two branches: (1) $6p$ parameters for a $p$-component Gaussian Mixture Model (GMM) that predicts the coordinate distribution of $(x,y)$, and (2) $2$ logits for classifying the pen state, i.e., \textit{Pen Down} or \textit{Pen Up}. 

\vspace{0.5mm} 
\textbf{Reconstruction Objective.}
The reconstruction objective consists of a coordinate loss $\mathcal{L}_{\mathrm{gmm}}$ and a pen-state loss $\mathcal{L}_{\mathrm{pen}}$. Given the ground-truth point $\mathbf{p}_t=(x_t,y_t)$, the coordinate loss is defined as the negative log-likelihood under the predicted GMM. 
As shown in \eqref{eq:gmm_loss}, where $\Omega$ denotes the set of valid non-padded trajectory steps, for each mixture component $m$, the GMM branch predicts the mixing coefficient $\pi_{m,t}$, mean $\boldsymbol{\mu}_{m,t}$, and covariance $\boldsymbol{\Sigma}_{m,t}$.
\begin{equation}
    \mathcal{L}_{\mathrm{gmm}} =
    -\frac{1}{|\Omega|}\sum_{t \in \Omega}
    \log \left( \sum_{m=1}^{p} \pi_{m,t}
    \mathcal{N}(\mathbf{p}_t \mid \boldsymbol{\mu}_{m,t}, \boldsymbol{\Sigma}_{m,t}) \right),
    \label{eq:gmm_loss}
\end{equation}
The pen-state branch is optimized with a focal loss~\citep{lin2017focal} defined as \eqref{eq:pen_loss}, where $y_{t,k}^{\mathrm{pen}}$ and $p_{t,k}^{\mathrm{pen}}$ denote the ground-truth label and predicted probability of pen-state class $k$, respectively.
\begin{equation}
    \mathcal{L}_{\mathrm{pen}} =
    -\frac{1}{N}\sum_{t=1}^{N}\sum_{k=1}^{2}
    \alpha_k (1-p_{t,k}^{\mathrm{pen}})^{\gamma}
    y_{t,k}^{\mathrm{pen}}\log p_{t,k}^{\mathrm{pen}},
    \label{eq:pen_loss}
\end{equation}
    
The final reconstruction loss is defined as \eqref{eq:rec_loss},  We compute $\mathcal{L}_{\mathrm{gmm}}$ over valid trajectory steps, while $\mathcal{L}_{\mathrm{pen}}$ is applied across the entire sequence, including padded regions, enabling the decoder to learn pen transitions and stopping behavior.
\begin{equation}
    \mathcal{L}_{\mathrm{rec}} = \mathcal{L}_{\mathrm{gmm}} + \mathcal{L}_{\mathrm{pen}}.
    \label{eq:rec_loss}
\end{equation}

\vspace{1mm} 
\textbf{Symbol- and Graph-Aware Regularization.}
To further align the latent space with mathematical structures, we introduce a task-aware perceptual loss $\mathcal{L}_{\mathrm{per}}$ as a structural regularizer. This loss consists of two complementary components: a symbol-level perceptual loss and a structure-level perceptual loss. Although they focus on different aspects of handwritten mathematical expressions, symbols and structures are inherently coupled: structural relations are defined over symbols, and mathematical structures become meaningful only when instantiated with concrete symbolic content. Therefore, we integrate symbol perception and relation perception into a unified OCR-based recognition framework, encouraging the encoder to learn representations that are both semantically discriminative and topology-aware.

Specifically, we employ a Transformer-based OCR module to predict an interleaved structural sequence $[R_\text{root}, S_1, R_2, S_2, \dots, R_n, S_n]$, which is obtained by linearizing the RelAST representation. Here, $S_i$ denotes the $i$-th symbol token, and $R_i$ represents its associated spatial relation.
The predicted probability sequence $\mathbf{P}^{\mathrm{ocr}}$ is optimized against the ground-truth structural sequence $\mathbf{y}^{\mathrm{ocr}}$ using Connectionist Temporal Classification (CTC~\citep{graves2006connectionist}) loss. This objective jointly enforces symbol-level recognition and relation-level topology consistency:
\begin{equation}
    \mathcal{L}_{\mathrm{per}} = \mathcal{L}_{\mathrm{sym}/{\mathrm{rel}}}
    = 
    \mathrm{CTC} \left( \mathbf{P}^{\mathrm{ocr}}, \mathbf{y}^{\mathrm{ocr}} \right).
    \label{eq:ocr_ctc}
\end{equation}
By optimizing $\mathcal{L}_{\mathrm{sym}}$ and $\mathcal{L}_{\mathrm{rel}}$ within a single sequence prediction objective, the encoder is encouraged to perceive a formula as a cohesive structural entity, leading to latent representations that better preserve both symbol identities and spatial layouts.

\vspace{1mm} 
\textbf{VAE Loss Function.}
MathVAE is trained end-to-end with the objective function $\mathcal{L}_{\mathrm{vae}}$ defined in Eq.~\ref{eq:vae_loss}. Besides reconstruction and perceptual regularization, we include a KL divergence term following the variational autoencoder formulation~\citep{kingma2013auto}, which regularizes the latent distribution toward a standard Gaussian prior. This enables the encoder to learn structurally coherent and semantically aligned latent representations for diffusion-based handwriting synthesis.
{
    \begin{equation}
    \mathcal{L}_{\text{vae}} = \sum\nolimits \lambda_\ell\cdot \mathcal{L}_\ell 
    \quad \text{where } \ell \in \{\mathrm{rec},\, \mathrm{kl},\, \mathrm{per}\}
    \label{eq:vae_loss}
    \end{equation}
}

\subsection{HMEG via MathDiT}
\label{sec:dit}

As shown in Fig.~\ref{fig_model_vae_dit} (c), MathDiT employs a Transformer-based diffusion architecture for HMEG within the latent space. Built upon the compact and structured manifold of MathVAE, MathDiT progressively refines noisy latent variables into coherent stroke trajectories through an iterative denoising process. Unlike traditional models conditioned on raw LaTeX strings, MathDiT is conditioned on the proposed RelAST, which provides an explicit structural prior through its serialized triplets $[S, R, D]$. By leveraging this topology-aware input, MathDiT achieves precise and structural generation, ensuring high fidelity and superior readability even for complex, deeply nested mathematical expressions.

\vspace{1mm} 
\textbf{Structural Text Representation.}
Compared with raw LaTeX input, we explicitly model symbols and relational nodes to capture spatial structures. By parsing each LaTeX expression into RelAST, we obtain its triplet representation $[S, R, D]$, which serves as the structured textual sequence.
Specifically, symbols $S$ and relations $R$ are projected into high-dimensional latent spaces via separate embedding layers. To capture the recursive hierarchy of mathematical notation, the nesting depth $D$ is encoded as a depth embedding. This depth information is integrated with its corresponding symbol embedding $S$ through element-wise addition, whereas relation nodes remain depth-invariant. The resulting sequence follows an interleaved structural pattern: $[R_{\text{root}}, S_1+D_1, R_2, S_2+D_2, \dots]$, originating from a dedicated $R_{\text{root}}$ node. This selective integration of depth-aware embeddings allows the model to anchor symbols within their respective hierarchical levels while using relations as topological bridges, strengthening its capacity for modeling complex, multi-scale mathematical structures.

\vspace{1mm} \noindent
\textbf{Complexity-Sensitive Symbol Count Prior.}
Long mathematical expressions are more challenging to generate due to their increased structural complexity. We therefore use the number of entity symbols in RelAST as a global complexity prior and incorporate it into the diffusion process through a \textit{Complexity-Sensitive Time Modulation} mechanism.

Let $N_c$ denote the number of entity symbols derived from RelAST. Instead of directly concatenating this signal with the structural condition, we adaptively rescale the timestep embedding according to the symbol count:
\vspace{-0.5mm}
\begin{equation}
\hat{\mathbf{t}} = \mathrm{MLP}(\mathbf{t})
\odot
\left(1+\tanh(\alpha)\left(1-{t}/{T}\right)\log(1+N_c)\right),
\label{eq:time_mod}
\end{equation}
where $\alpha$ is a learnable scalar, $t$ is the current diffusion timestep, and $T$ is the total number of diffusion steps. The logarithmic mapping stabilizes the symbol-count signal, while $\left(1-\frac{t}{T}\right)$ gradually increases its influence during denoising. As a result, the model first recovers the global layout and progressively incorporates expression complexity during later refinement, reducing symbol omission and over-generation in complex formulas.

\vspace{1mm} 
\textbf{Diffusion Modeling.}
MathDiT performs conditional denoising in the structured latent space learned by MathVAE. During training, a clean latent representation $x_0$ is first perturbed into a noisy latent $x_t$ according to the diffusion schedule:
\begin{equation}
x_t = \sqrt{\bar{\alpha}_t}\,x_0 
+ \sqrt{1-\bar{\alpha}_t}\,\epsilon,
\qquad 
\epsilon \sim \mathcal{N}(\mathbf{0}, \mathbf{I}).
\label{equation:x_t}
\end{equation}

To incorporate structural guidance, the noisy latent $x_t$ is concatenated with the RelAST condition $Z$ along the channel dimension, and then projected to the hidden dimension of the Transformer. Given the timestep $t$ and the symbol count $N_c$, MathDiT directly predicts the clean latent representation:
\begin{equation}
\hat{x}_0 = \mathrm{MathDiT}_{\theta}([x_t, Z], t, N_c).
\label{equation:dit_pred}
\end{equation}

The model is optimized by minimizing the reconstruction error between the predicted clean latent $\hat{x}_0$ and the ground-truth clean latent $x_0$:
\begin{equation}
\mathcal{L}_{\mathrm{diff}}
=
\mathbb{E}_{x_0,t,\epsilon}
\left[
\left\|
\hat{x}_0 - x_0
\right\|_2^2
\right].
\label{equation:loss_diff}
\end{equation}

Within the denoising blocks, the complexity-aware embedding $\hat{t}$ is injected via Adaptive Layer Normalization (AdaLN)~\citep{DiT}, enabling the refinement process to be dynamically modulated by global symbol density. Together with the RelAST condition $Z$, this design guides the denoising trajectory to satisfy both structural constraints and global complexity priors, yielding high-fidelity and topologically consistent expressions.

During inference, generation starts from Gaussian noise $x_T \sim \mathcal{N}(\mathbf{0}, \mathbf{I})$. MathDiT iteratively predicts the clean latent representation conditioned on $Z$ and $N_c$, and progressively refines the latent through the reverse diffusion process. The final denoised latent is decoded by MathVAE into an online handwritten mathematical expression.

\section{Experiments}
\label{sec:experiments}

\subsection{Experimental Setup}
\label{sec:dataset_metrics}

\noindent
\textbf{Data Preparation.}
We utilize the recent and relatively challenging MathWriting~\citep{philippe2025mathwriting} dataset for HMEG. Our experimental setup uses 220k authentic human-written samples for training and a dedicated test set of over 7k samples. By parsing LaTeX into RelAST, we obtain a vocabulary of 244 tokens, including digits, Latin/Greek characters, operators, and relational symbols.

\vspace{1mm} 
\textbf{Baseline Methods.}
\label{sec:sota_compare}

We select several domain-relevant generative methods, including FormulaGAN~\citep{formula_gan}, a GAN-based approach, as well as One-DM~\citep{OneDM} and DiffInk~\citep{pan2026diffink}, which serve as strong baselines for handwritten content generation. We also include high-capacity text-to-image models, including SD-XL~\citep{podell2024sdxl}, FLUX.1~\citep{labs2025flux1kontextflowmatching}, Z-Image~\citep{team2025zimage}, and Qwen-Image~\citep{wu2025qwenimagetechnicalreport}, to assess generation quality.

\vspace{1mm} 
\textbf{Evaluation.}

HMEG requires both content fidelity and handwriting realism. We evaluate content fidelity using the state-of-the-art (SOTA) HMER model Uni-MuMER~\citep{li2025unimumer} with Edit, Expression Recognition Rate (ExpRate), and Bilingual Evaluation Understudy (BLEU). Handwriting realism is assessed by Fréchet Inception Distance (FID)~\citep{FID} and a user study on content correctness, structural correctness, and handwriting style.

\vspace{1mm} \noindent
\textbf{Implementation Details.}
MathVAE is trained using the AdamW optimizer with an initial learning rate of $1.5\times10^{-4}$, while MathDiT uses AdamW with an initial learning rate of $3\times10^{-4}$. The diffusion model employs $T=1000$ timesteps with a cosine noise schedule. During training, classifier-free guidance (CFG~\citep{ho2021classifierfree}) is enabled by randomly dropping the conditioning input with a probability of 10\%. During inference, the guidance scale is set to 3.0, and DDIM~\citep{song2021ddim} sampling with 20 steps is adopted for efficient generation.

\begin{table}[t]
    \centering
    \scriptsize
    \setlength{\tabcolsep}{6pt}
    \renewcommand{\arraystretch}{1.05}
    \caption{
    \textbf{Quantitative comparison with baseline methods on the MathWriting benchmark.}
    }
    \label{tab:compare_mathwriting}
    \begin{tabular}{c c c c c c c}
    \toprule
    \textbf{Methods}
    & \textbf{Venue}
    & \textbf{Edit.\%~$\uparrow$}
    & \textbf{Exp.\%~$\uparrow$}
    & \textbf{BLEU~$\uparrow$}
    & \textbf{FID~$\downarrow$}
    & \textbf{User\%~$\uparrow$} \\
    \midrule

    FormulaGAN (Springstein et al.,~\citeyear{formula_gan})
    & ICMR'21 & 95.06 & 65.86 & 92.14 & 98.15 & 2.6 \\

    One-DM~\citep{OneDM}
    & ECCV'24 & 88.20 & 39.66 & 82.02 & 29.33 & 4.1 \\

    SD-XL~\citep{podell2024sdxl}
    & ICLR'24 & 88.93 & 35.00 & 81.39 & 9.50 & 7.9 \\

    FLUX.1~\citep{labs2025flux1kontextflowmatching}
    & arXiv'25 & 91.06 & 53.12 & 85.21 & 16.06 & 10.4 \\

    Z-Image~\citep{team2025zimage}
    & arXiv'25 & 78.82 & 22.21 & 61.51 & 8.04 & 5.3 \\

    Qwen-Image~\citep{wu2025qwenimagetechnicalreport}
    & arXiv'25 & 94.63 & 65.37 & 91.83 & 10.32 & 21.4 \\

    DiffInk~\citep{pan2026diffink}
    & ICLR'26 & 92.33 & 58.35 & 90.15 & 8.24 & 17.0 \\

    \textbf{DiffMath}
    & \textbf{This Work}
    & \textbf{95.47} & \textbf{70.70} & \textbf{93.05} & \textbf{5.43} & \textbf{31.3} \\

    \bottomrule
    \end{tabular}
\end{table}

\begin{figure} [t]
    \centering
    \includegraphics[width=1\linewidth]{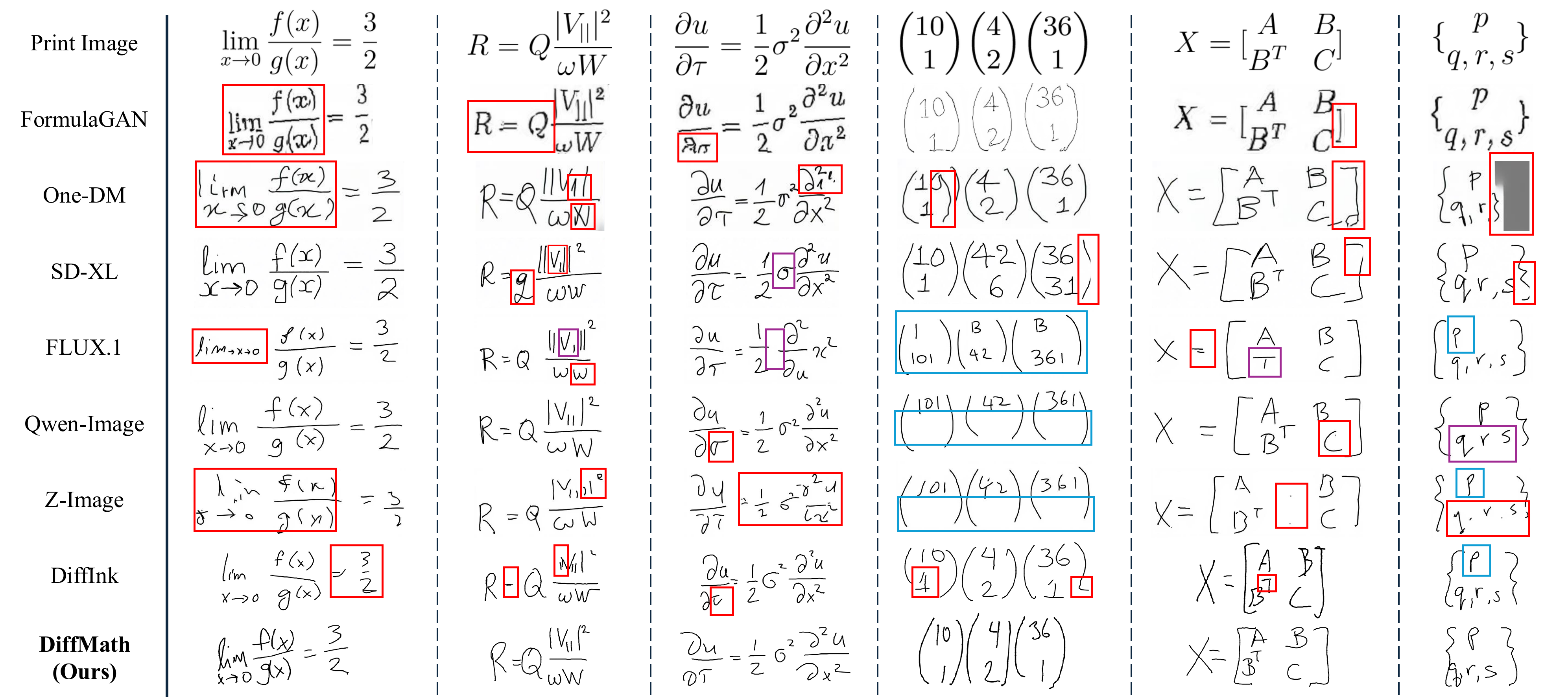}
    \caption{
    \textbf{Qualitative comparison results.}
    Red boxes highlight atomic content errors (incorrect symbols or artifacts), blue boxes indicate structural misalignments, and purple boxes denote missing symbols. Please zoom in for a better view of the details. }
    \label{fig:main_results}
\end{figure}

\subsection{Main Results}
\textbf{Quantitative Evaluation.}
As shown in Tab.~\ref{tab:compare_mathwriting}, for content fidelity, DiffMath (Ours) obtains an ExpRate of 70.70\%, outperforming FormulaGAN by 4.84\% and exceeding SD-XL (35.00\%) and FLUX.1 (53.12\%). These results suggest that the proposed RelAST representation provides more effective guidance than raw LaTex input. DiffMath also achieves the lowest FID score (5.43), indicating that the generated samples are closer to the real data distribution. Although large-scale models such as Qwen-Image produce visually plausible results, their structural accuracy remains lower. Compared with DiffInk, a method for unstructured handwritten text-line generation, DiffMath improves ExpRate by 12.35\%, highlighting the advantage of explicit structural modeling. This trend is further supported by the user study, where DiffMath receives the highest preference score.

\vspace{1mm} 
\textbf{Qualitative evaluation.}
As shown in Fig.~\ref{fig:main_results}, red and blue boxes highlight content and structural errors, respectively. While FormulaGAN struggle with realistic handwriting styles, more recent models like One-DM often suffer from background artifacts and character distortions. Even powerful pre-trained models (e.g., FLUX.1, Qwen-Image) frequently fail in dense symbol scenarios like matrices. Similarly, DiffInk exhibits structural collapse due to a lack of explicit layout modeling. In contrast, DiffMath maintains high-fidelity generation and robust structural integrity. 
Additional comparison results and generated samples can be found in Fig.~\ref{fig:more_results} and Fig.~\ref{fig:more_results_self}.

\subsection{Ablation Study of DiffMath}

\noindent\textbf{Quantitative analysis.}
Tab.~\ref{tab:ablation} reports the quantitative ablation results of DiffMath. For MathVAE, introducing the symbol-aware regularization \textbf{$\mathcal{L}_\text{sym}$} substantially improves the learned latent representation, increasing ExpRate by 19 points while reducing FID. Further incorporating the relation-aware regularization \textbf{$\mathcal{L}_\text{rel}$} yields additional gains, achieving an ExpRate of 70.70\% and an FID of 5.43. For MathDiT, progressively incorporating relation tokens \textbf{$\text{Rel.}$}, depth embeddings \textbf{$\text{Dep.}$}, and the global symbol-count prior \textbf{$\text{Glo.}$} consistently improves structural accuracy and generation quality, with the full model achieving the best performance across all evaluation metrics.

\begin{figure}[t]
    \centering
    \begin{minipage}[t]{0.48\textwidth}
        \centering
        \includegraphics[width=\linewidth]{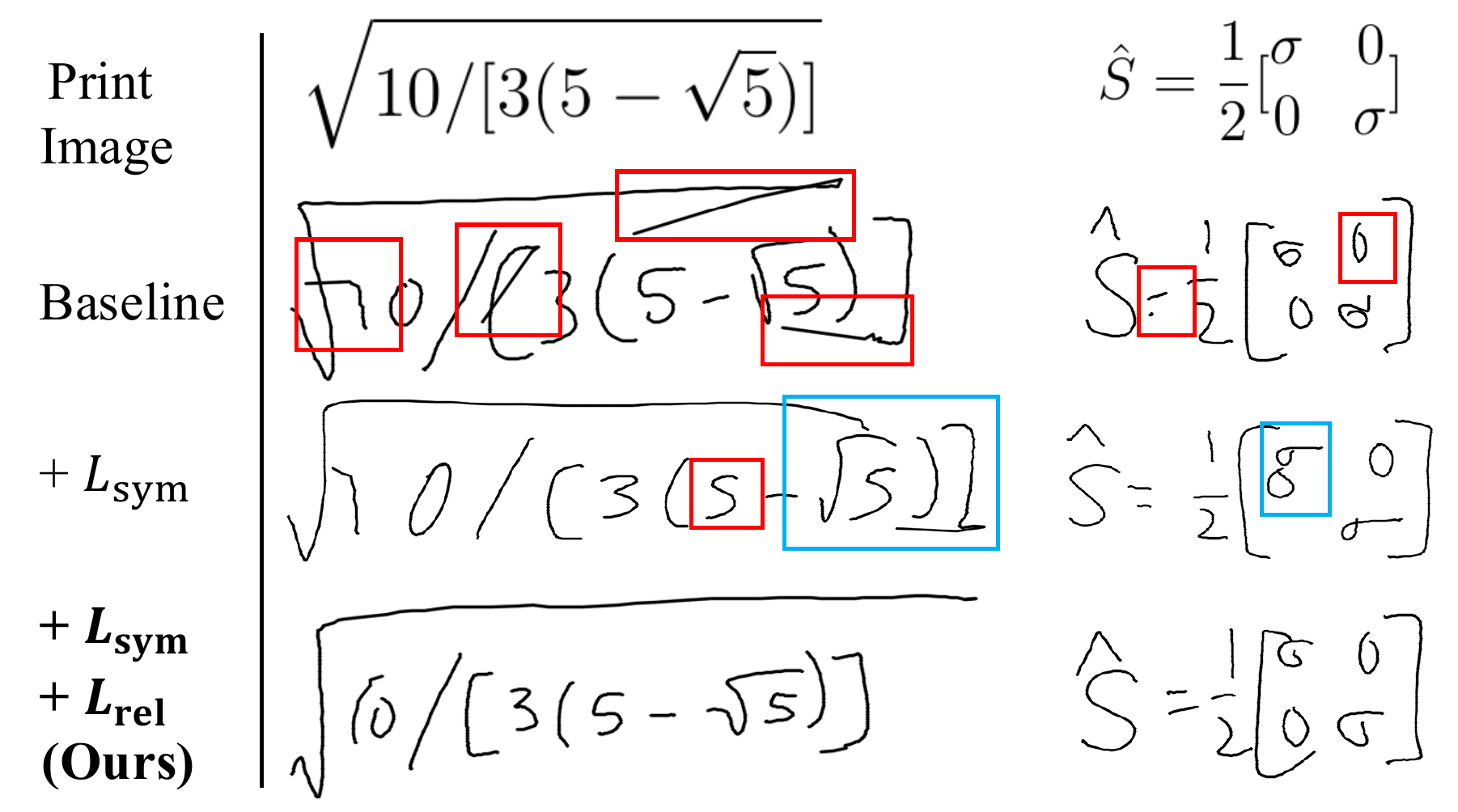}
    \end{minipage}
    \hfill
    \begin{minipage}[t]{0.48\textwidth}
        \centering
        \includegraphics[width=\linewidth]{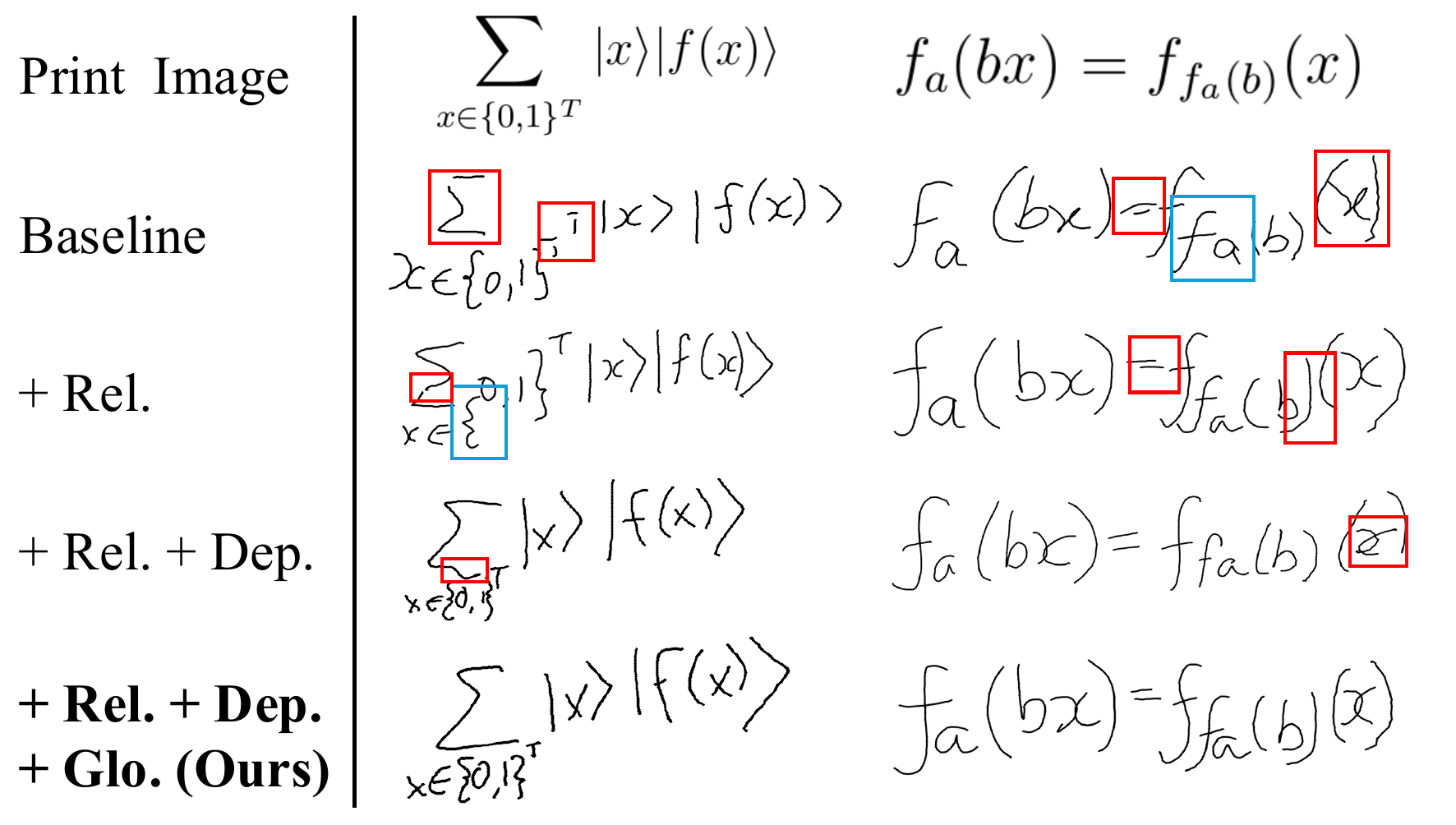}
    \end{minipage}

    \caption{\textbf{Visual ablation studies.}
    The left and right figures compare different MathVAE and MathDiT variants, respectively. Red and blue boxes highlight content and structural/style errors. Incorporating all proposed components yields the best visual quality and structural consistency.}
    \label{fig:vae_dit_ablation}
\end{figure}

\vspace{1mm}
\noindent\textbf{Visual analysis.}
Fig.~\ref{fig:vae_dit_ablation} provides qualitative comparisons of different DiffMath variants. The left figure shows that the proposed MathVAE improves symbol correctness and structural consistency by learning a more informative latent representation. The right figure demonstrates that progressively introducing structural priors into MathDiT effectively reduces content omissions and layout errors, resulting in more accurate and coherent handwritten mathematical expressions.

\begin{table}[t]
    \centering
    \scriptsize
    \setlength{\tabcolsep}{3pt}
    \renewcommand{\arraystretch}{1.0}
    
    \caption{\textbf{Quantitative ablation studies of DiffMath.} (a) Effect of MathVAE on latent representation learning. (b) Effect of structural priors in MathDiT.}
    \label{tab:ablation}
    \renewcommand{\arraystretch}{1.2}
    \begin{tabular*}{0.49\linewidth}{@{\extracolsep{\fill}}ccc|cccc@{}}
    \multicolumn{7}{c}{\textbf{(a) Effect of MathVAE}}\\
    \hline
    \multicolumn{3}{c|}{\textbf{VAE Loss}} & \multicolumn{4}{c}{\textbf{Generation Performance}}\\
    \hline
    $\mathcal{L}_{\rm rec+kl}$ & $\mathcal{L}_{\rm sym}$ & $\mathcal{L}_{\rm rel}$ & Edit & Exp & BLEU & FID\\
    \hline
    \cmark & & & 89.77 & 46.23 & 81.55 & 18.24\\
    \cmark & \cmark & & 92.75 & 65.22 & 91.55 & 6.56\\
    \cmark & \cmark & \cmark & \textbf{95.47} & \textbf{70.70} & \textbf{93.05} & \textbf{5.43}\\
    \hline
    \end{tabular*}
    \hfill
    \renewcommand{\arraystretch}{1.0}
    \begin{tabular*}{0.49\linewidth}{@{\extracolsep{\fill}}cccc|cccc@{}}
    \multicolumn{8}{c}{\textbf{(b) Effect of MathDiT}}\\
    \hline
    \multicolumn{4}{c|}{\textbf{Condition Inputs}} & \multicolumn{4}{c}{\textbf{Generation Performance}}\\
    \hline
    Sym. & Rel. & Dep. & Glo. & Edit & Exp & BLEU & FID\\
    \hline
    \checkmark & & & & 91.25 & 56.27 & 88.92 & 8.62\\
    \checkmark & \checkmark & & & 93.91 & 63.89 & 89.20 & 5.99\\
    \checkmark & \checkmark & \checkmark & & 95.46 & 69.83 & 92.65 & 5.44\\
    \checkmark & \checkmark & \checkmark & \checkmark & \textbf{95.47} & \textbf{70.70} & \textbf{93.05} & \textbf{5.43}\\
    \hline
    \end{tabular*}

\end{table}

\begin{wraptable}{r}{0.62\linewidth}
    \centering
    \scriptsize
    \vspace{-8pt}
    \setlength{\tabcolsep}{3pt}
    \renewcommand{\arraystretch}{1.25}
    \caption{\textbf{Downstream OCR experiments.}
    ``Aug.'' denotes conventional data augmentation, while ``Syn.'' denotes synthetic data generated by DiffMath (\textbf{Ours}). For ``Syn.'', we add 150k and 30k generated samples to the MathWriting and CROHME training sets, respectively.}
    \label{tab_hmeg4hmer}

    \resizebox{\linewidth}{!}{
    \begin{tabular}{clllll}
    \hline
    \multirow{2}{*}{\textbf{OCR Model}}
    & \multirow{2}{*}{\textbf{Training Data}}
    & \multicolumn{2}{c}{\textbf{MathWriting}}
    & \multicolumn{2}{c}{\makecell{\textbf{Avg. CROHME}\\\textbf{(14/16/19)}}} \\

    \cline{3-4}\cline{5-6}

    &
    & \textbf{Exp.\%}
    & \textbf{@CDM}
    & \textbf{Exp.\%}
    & \textbf{@CDM} \\

    \hline

    \multirow{3}{*}{\begin{tabular}[c]{@{}l@{}}
        \textbf{Qwen3-VL}\\
        \textbf{8B-Instruct}
    \end{tabular}}
    & Original Data
    & 67.22
    & 70.93
    & 69.19
    & 72.34 \\

    \cline{2-6}

    & + Aug. Data
    & 67.84\textsuperscript{\textcolor{blue}{+0.62}}
    & 71.05\textsuperscript{\textcolor{blue}{+0.12}}
    & 69.56\textsuperscript{\textcolor{blue}{+0.37}}
    & 72.37\textsuperscript{\textcolor{blue}{+0.03}} \\

    \cline{2-6}

    & \textbf{+ Syn. Data}
    & \textbf{68.58}\textsuperscript{\textcolor{red}{+1.36}}
    & \textbf{71.79}\textsuperscript{\textcolor{red}{+0.86}}
    & \textbf{72.40}\textsuperscript{\textcolor{red}{+3.21}}
    & \textbf{75.08}\textsuperscript{\textcolor{red}{+2.74}} \\

    \hline

    \multirow{3}{*}{\begin{tabular}[c]{@{}l@{}}
        \textbf{Uni-MuMER}\\
        \textbf{(Current SOTA)}
    \end{tabular}}
    & Original Data
    & 68.58
    & 74.11
    & 77.70
    & 80.88 \\

    \cline{2-6}

    & + Aug. Data
    & 70.01\textsuperscript{\textcolor{blue}{+1.43}}
    & 74.38\textsuperscript{\textcolor{blue}{+0.27}}
    & 78.05\textsuperscript{\textcolor{blue}{+0.35}}
    & 80.95\textsuperscript{\textcolor{blue}{+0.07}} \\

    \cline{2-6}

    & \textbf{+ Syn. Data}
    & \textbf{71.42}\textsuperscript{\textcolor{red}{+2.84}}
    & \textbf{75.01}\textsuperscript{\textcolor{red}{+0.90}}
    & \textbf{79.17}\textsuperscript{\textcolor{red}{+1.47}}
    & \textbf{81.82}\textsuperscript{\textcolor{red}{+0.94}} \\

    \hline
    \end{tabular}
    }
\end{wraptable}

\subsection{Data Augmentation for HMER}
HMEG provides an effective way to augment training data for HMER while reducing annotation costs. To evaluate DiffMath as a data generator, we compare its synthetic data with conventional augmentation on MathWriting~\citep{philippe2025mathwriting} and the CROHME~\citep{crohme14,crohme16,crohme19} benchmark series, using sequence-level ExpRate together with CDM~\citep{Wang_2025_CVPR} and ExpRate@CDM. As shown in Tab.~\ref{tab_hmeg4hmer}, we adopt a strong augmentation recipe from prior HMER studies~\citep{guan2024posformer}, yet it provides only marginal gains, whereas DiffMath-generated samples consistently yield larger improvements across Qwen3-VL-8B-Instruct~\citep{qwen3technicalreport} and the SOTA Uni-MuMER~\citep{li2025unimumer}. 
This difference stems from the fact that conventional augmentation mainly perturbs existing samples and thus preserves the underlying expression distribution, while generative augmentation can leverage LaTeX expressions beyond the original training set to introduce new symbol combinations and spatial structures. This capability effectively expands both the quantity and structural diversity of training data, which is particularly valuable under the current data-scaling paradigm for improving recognition performance.

\vspace{-2pt}
\section{Conclusion}
\label{sec:conclusion}

In this paper, we presented DiffMath, a structure-aware latent diffusion framework for HMEG that derives structural priors from LaTeX without requiring layout annotations or symbol-level positional supervision. By introducing RelAST, DiffMath explicitly models symbol identities, two-dimensional spatial relations, and hierarchical layouts, providing informative structural conditions for handwritten mathematical expression generation. Built upon this representation, MathVAE learns a structure-preserving latent space, while MathDiT performs structure-conditioned latent diffusion to generate handwritten mathematical expressions with high visual fidelity and structural consistency.
Extensive experiments demonstrated that DiffMath consistently outperformed existing methods, particularly in preserving inter-symbol relations, superscript and subscript structures, and global expression layouts. Furthermore, the generated samples served as effective synthetic data for HMER, consistently improving multiple recognition models across different benchmarks. Notably, these gains were larger than those obtained with conventional data augmentation, suggesting the potential of generative data augmentation for enriching HMER training data. These results provide a perspective on structured generative modeling by highlighting the value of explicit structural priors.










\bibliography{iclr2027_conference}
\bibliographystyle{iclr2027_conference}

\newpage
\appendix
\section{The Proposed DiffMath Framework}

\subsection{RelAST Construction.}

Given a LaTeX expression, we first convert it into Presentation MathML using the LaTeXML~\cite{latexml} toolkit, which exposes the structural hierarchy of mathematical expressions in a tree representation. Based on the resulting MathML structure, we recursively construct a relation-aware abstract syntax tree (RelAST), as described in Algorithm~\ref{alg:mathml2relast}. Atomic symbols are mapped to leaf nodes, while sequential layouts are connected using the \textit{right} relation to preserve reading order. Two-dimensional mathematical structures are explicitly represented by spatial relations, including \textit{sup}, \textit{sub}, \textit{above}, \textit{below}, and \textit{inside}, corresponding to superscripts, subscripts, fractions, and roots. For matrix-like expressions, hierarchical row–cell structures are introduced to capture the two-dimensional layout.

To obtain a sequence representation suitable for generative modeling, the parsed structure is linearized via a depth-first traversal. Each node is represented as a tuple $[S, R, D]$, where $S$ denotes the symbol, $R$ the incoming relation, and $D$ the tree depth. The resulting symbol–relation–depth sequence is defined as the RelAST representation, which preserves both symbol identities and structural dependencies of the original formula while providing a compact sequential format.

\begin{Algorithm}
    \caption{LaTeX to RelAST Conversion}
    \label{alg:mathml2relast}
    
    \begin{algorithmic}[1]
    
    \Require LaTeX expression $L$
    \Ensure RelAST node $T$
    
    \State $M \gets \textsc{LaTeXML}(L)$ \Comment{convert LaTeX to Presentation MathML}
    
    \Function{BuildRelAST}{$x$}
    
    \State $tag \gets \textsc{Tag}(x)$
    
    \If{$tag$ is a symbol node ($mi$, $mn$, $mo$, $mtext$)}
        \State \Return \textsc{CreateSymbolNode}$(x)$
    \EndIf
    
    \If{$tag$ denotes a linear container ($mrow$)}
        \State $C \gets [\textsc{BuildRelAST}(c)\ |\ c \in Children(x)]$
        \State \Return \textsc{ConnectSequentially}$(C, right)$
    \EndIf
    
    \If{$tag$ denotes a script structure ($msub$, $msup$, $msubsup$)}
        \State $base \gets \textsc{BuildRelAST}(Base(x))$
        \For{each script $s$ in $x$}
            \State \textsc{Attach}$(base,\ \textsc{BuildRelAST}(s),\ ScriptRelation(s))$
        \EndFor
        \State \Return $base$
    \EndIf
    
    \If{$tag$ denotes a fraction or root}
        \State $T \gets \textsc{CreateNode}(tag)$
        \For{each child $c$ in $x$}
            \State \textsc{Attach}$(T,\ \textsc{BuildRelAST}(c),\ StructuralRelation(c))$
        \EndFor
        \State \Return $T$
    \EndIf
    
    \If{$tag = mtable$}
        \State \Return \textsc{BuildTableTree}(x)
    \EndIf
    
    \State $C \gets [\textsc{BuildRelAST}(c)\ |\ c \in Children(x)]$
    \State \Return \textsc{ConnectSequentially}$(C, right)$
    
    \EndFunction
    
    \State \Return \textsc{BuildRelAST}$(Root(M))$
    
    \end{algorithmic}
\end{Algorithm}

\begin{figure}[t]
    \centering
    \includegraphics[width=1\linewidth]{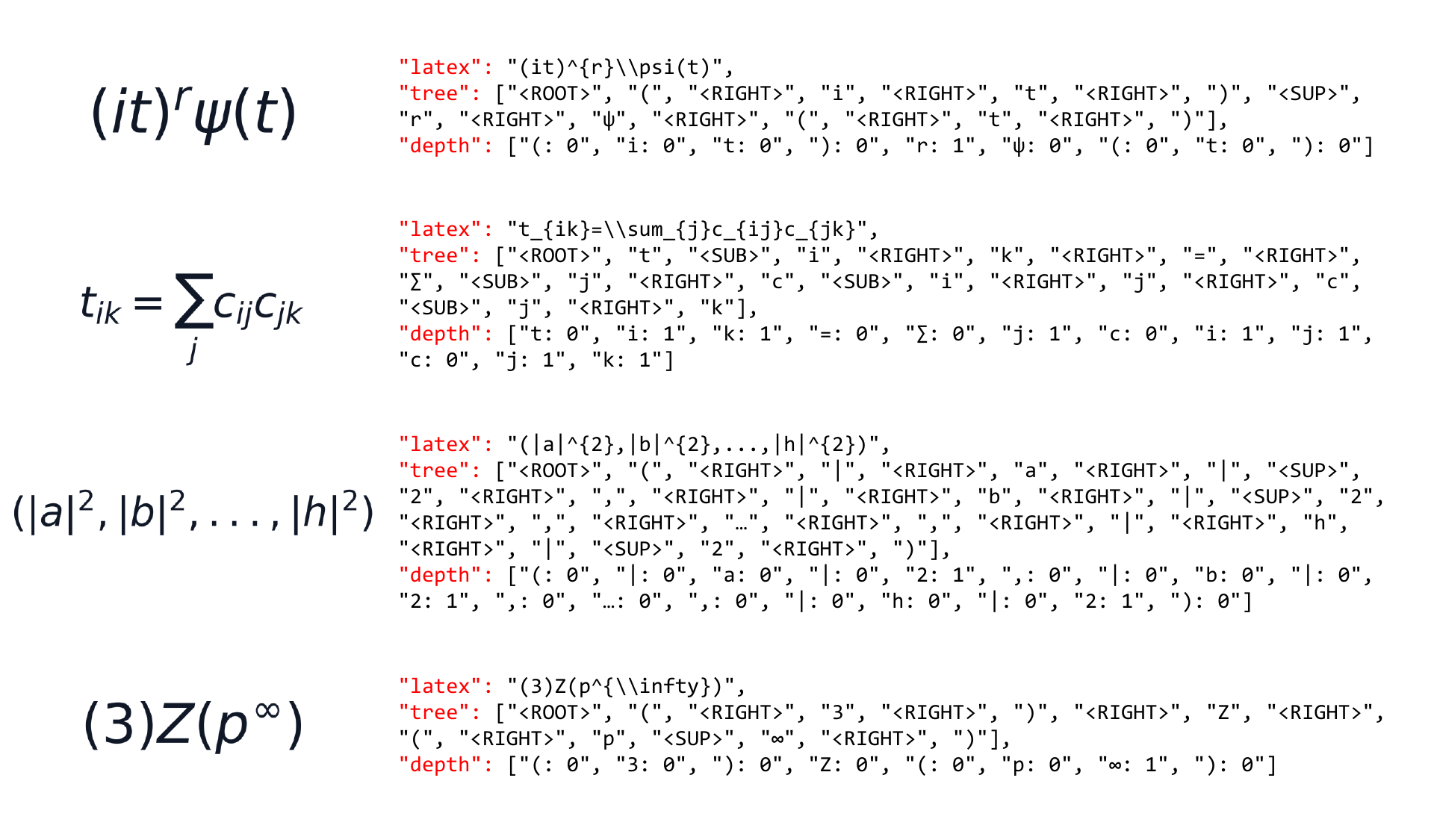}
    \caption{
    Examples of RelAST representations, illustrating how mathematical expressions are organized into compact structural descriptions.
    }
    \label{fig:RelAST}
\end{figure}

\subsection{MathVAE Architecture}

MathVAE serves as the latent representation module for the latent diffusion Transformer, learning compact latent features while preserving reconstruction fidelity. It adopts a hierarchical 1D convolutional encoder–decoder with residual connections. The encoder consists of two strided convolutional layers, each followed by a four-layer residual stack, progressively compressing the input trajectory from $\mathbb{R}^{N\times4}$ to a latent representation of $\mathbb{R}^{N/4\times384}$. The decoder mirrors this architecture using residual stacks and transposed convolutions to progressively recover the temporal resolution. The reconstructed features are further processed by a Transformer-based trajectory decoder comprising a linear projection, a 4-layer Transformer encoder (hidden size 256, 4 attention heads), and an output projection, which predicts the parameters of a $p=20$ component Gaussian Mixture Model (GMM~\cite{reynolds2015gaussian}) for $(x,y)$ trajectory generation, while the remaining dimensions are used for pen-state classification. To preserve semantic information in the latent space, we additionally employ an auxiliary OCR branch on the encoder features. It consists of a linear projection, sinusoidal positional encoding, and a 3-layer Transformer encoder (hidden size 256, 4 attention heads), followed by a linear classifier optimized with the CTC loss, encouraging the latent representations to retain character-discriminative information without explicit point-level alignment.

\subsection{MathDiT Architecture}

The overall training and inference procedures are summarized in Algorithm~\ref{alg:dit_training_inference}. MathDiT is trained to predict the clean latent representation $x_0$ from its noisy counterpart $x_t$, while inference progressively transforms Gaussian noise into the final latent sample. In implementation, MathDiT takes the noisy latent sequence, the RelAST-derived symbol and depth sequences, and the global symbol count $N_c$ as conditions. Symbol and depth embeddings are first summed to form 512-dimensional token representations and contextualized by three stacked ConvNeXt-V2\cite{woo2023convnext} blocks, each consisting of a depthwise 1D convolution, Layer Normalization, a two-layer feedforward network with GELU activation and Global Response Normalization (GRN), and residual connections. The resulting structure-aware content features are concatenated with the latent input and projected to a 768-dimensional representation, followed by a convolutional positional embedding. The fused features are then processed by a 16-layer DiT backbone with multi-head self-attention, feedforward networks, rotary positional embeddings, and AdaLN-Zero modulation. To incorporate global expression complexity, the symbol count $N_c$ is logarithmically normalized and used to modulate the magnitude of the timestep embedding through a learnable time-dependent scaling factor, providing stronger guidance during later denoising stages. Finally, the modulated features are projected back to the latent space to predict the clean latent representation.

\begin{Algorithm}[t]
    \caption{MathDiT Training and Inference}
    \footnotesize
    \vspace{0.5em}
    \begin{minipage}[t]{0.48\linewidth}
    \textbf{Training Phase: }
    \begin{algorithmic}[1]
      \State \textbf{Input:} $x_0,Z,N_c$
      \State Sample $t \sim \mathcal{U}(1, T)$
      \State Sample noise $\epsilon \sim \mathcal{N}(0, \mathbf{I})$
      \State $x_t = \sqrt{\bar{\alpha}_t} \cdot x_0 + \sqrt{1 - \bar{\alpha}_t} \cdot \epsilon$
      \State $\hat{x}_0 = \text{MathDiT}_\theta([x_t, Z],t,N_c)$
      \State Update $\text{MathDiT}_\theta$: $\|\hat{x}_0-x_0\|^2$.
    \end{algorithmic}
    \end{minipage}
    \hfill
    \begin{minipage}[t]{0.48\linewidth}
    \textbf{Inference Phase: }
    \begin{algorithmic}[1]
    \State \textbf{Input:} $x_T \sim \mathcal{N}(0, \mathbf{I}),Z,N_c$
    \For{$t = T, \dots, 1$}
      \State $\hat{x}_0 = \text{MathDiT}([x_t, Z],t,N_c)$
      \State $\epsilon = (x_t - \sqrt{\bar{\alpha}_t} \cdot \hat{x}_0) / \sqrt{1 - \bar{\alpha}_t}$
      \State $x_{t-1} = \sqrt{\bar{\alpha}_{t-1}} \cdot \hat{x}_0 + \sqrt{1 - \bar{\alpha}_{t-1}} \cdot \epsilon$
    \EndFor \Comment{Final output: $\hat{x}_0$}
    \end{algorithmic}
    \end{minipage}
    \label{alg:dit_training_inference}
\end{Algorithm}

\section{Implementation Details}

\subsection{Proposed DiffMath}

DiffMath is implemented using the PyTorch framework. 
Specifically, \textbf{MathVAE} is trained on a single NVIDIA H800 GPU with a batch size of 128 for 100 epochs. The training uses FP32 precision and the AdamW optimizer. The initial learning rate is set to $1.5\times10^{-4}$ with a minimum learning rate of $1\times10^{-5}$, and a warmup schedule of 5,000 steps is adopted. In the loss formulation, the weights of $L_{\text{gmm}}$, $L_{\text{pen}}$, and $L_{\text{per}}$ are set to 0.1, 1.0, and 1.0, respectively.

\textbf{MathDiT} is trained on 8 NVIDIA H800 GPUs with a per-GPU batch size of 128 for approximately 220k training steps. The model also uses FP32 precision and the AdamW optimizer. The initial learning rate is set to $3\times10^{-4}$ with a minimum learning rate of $3\times10^{-6}$, together with a warmup of 20k steps. The diffusion process uses 1000 timesteps with a cosine noise schedule. During training, the conditioning input is randomly dropped with a probability of 10\% to enable classifier-free guidance (CFG~\cite{ho2021classifierfree}). During inference, the guidance scale is set to 3.0. DDIM~\cite{song2021ddim} sampling is adopted for acceleration, with the number of sampling steps set to 20.

\subsection{Baseline Models}
For comparison, we include several representative generation models, including Formulagan~\cite{formula_gan}, One-DM~\cite{OneDM}, SD-XL~\cite{podell2024sdxl}, FLUX.1~\cite{flux-2-2025}, Z-Image~\cite{team2025zimage}, Qwen-Image~\cite{wu2025qwenimagetechnicalreport}, and DiffInk~\cite{pan2026diffink}. Among them, DiffInk, similar to our method, performs generation in the online sequence modality. Formulagan and One-DM are specialized models designed for formula generation, while SD-XL, FLUX.1, Z-Image, and Qwen-Image are general text-to-image generation models. Below we describe the training and inference details.

For the early GAN-based generation method FormulaGAN, we directly use the pretrained weights released by the authors for inference. For the handwritten text-line generation method DiffInk, we replace the original text input with LaTeX expressions as the conditioning input and remove the style branch in the VAE module, retaining only the content-related regularization. The remaining training and testing settings follow the default configuration. During inference, the CFG scale is set to 3.0 with 20 sampling steps.

For the dedicated handwriting generation model One-DM, we remove the style branch from the original method while keeping the remaining settings the same as the default configuration during training. During inference, we set the CFG scale to 4 to achieve better generation quality. 

For general-purpose text-to-image models, including Qwen-Image, Z-Image, SD-XL, and FLUX.1, we fine-tune the models using LoRA with rank = 64 for 5 epochs. During inference, the CFG scale is set to 5 for Qwen-Image and Z-Image, 5.0 for FLUX.1, and 7.5 for SD-XL to obtain better generation results.

\subsection{OCR Models}
For the OCR models, we adopt Qwen-3VL-8B Instruct~\cite{qwen3technicalreport} and Uni-Mumer~\cite{li2025unimumer} (built upon Qwen-2.5VL-3B) as the backbone models. When training with the original data as well as the dataset augmented with DiffMath-generated samples, we perform full-parameter fine-tuning. For both the MathWriting~\cite{philippe2025mathwriting} and CROHME~\cite{crohme14,crohme16,crohme19} datasets, the models are fine-tuned for one epoch.

\subsection{Details of Metrics}

For evaluation, we adopt four metrics: Expression Recognition Rate (ExpRate), Character Detection Matching (CDM), ExpRate@CDM (@CDM~\cite{Wang_2025_CVPR}), and a User Study. ExpRate measures the string-level accuracy between the predicted and ground-truth LaTeX expressions. 
Since string matching alone cannot fully reflect visual structural correctness, following Uni-Mumer~\cite{li2025unimumer}, we further adopt a more robust visual metric, CDM, which renders LaTeX expressions into images and evaluates consistency by detecting symbols and comparing their spatial layouts. ExpRate@CDM is a stricter criterion that requires the rendered results to be visually identical, reducing errors caused by different writing styles or equivalent layouts.
In addition, we conduct a user study for subjective evaluation. We randomly sample 200 LaTeX expressions from the test set and generate the corresponding handwritten formulas using each competing method. Fifteen volunteers are invited to participate in the evaluation. For each expression, participants are presented with the generated results from all methods and asked to select the one they consider the best, taking into account both the correctness of symbols and structural layout as well as the naturalness of the handwriting style. The user study score is then computed as the proportion of votes received by each method across all evaluations.

\section{More Experimental Results}
\subsection{Comparison with Baseline Methods.}
\begin{figure}[t]
\centering
    \begin{subfigure}{1\linewidth}
        \centering
        \includegraphics[width=\linewidth]{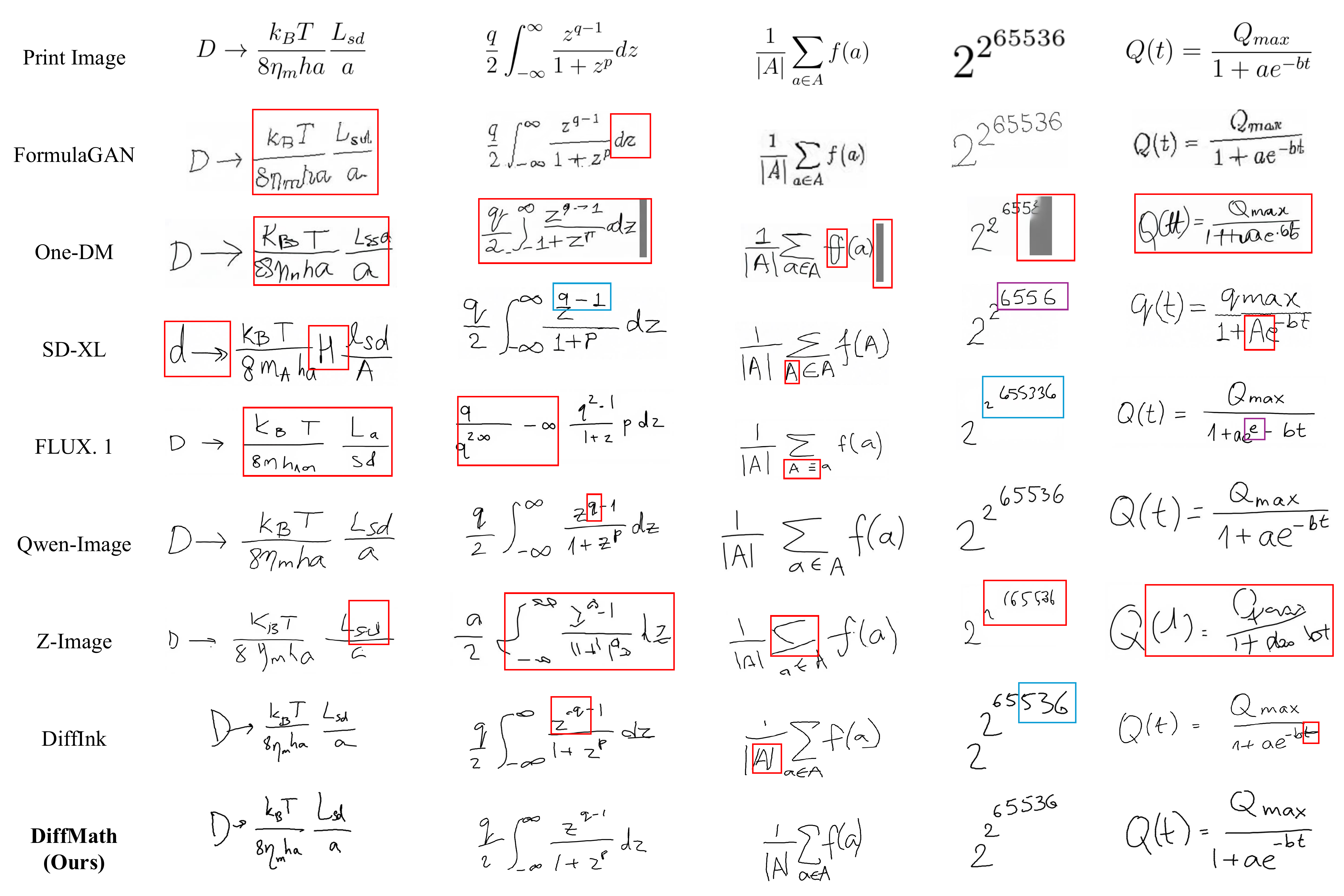}
        \caption{Comparison on \textbf{simple and common} expressions.}
    \end{subfigure}
    
    \vspace{3mm}
    
    \begin{subfigure}{1\linewidth}
        \centering
        \includegraphics[width=\linewidth]{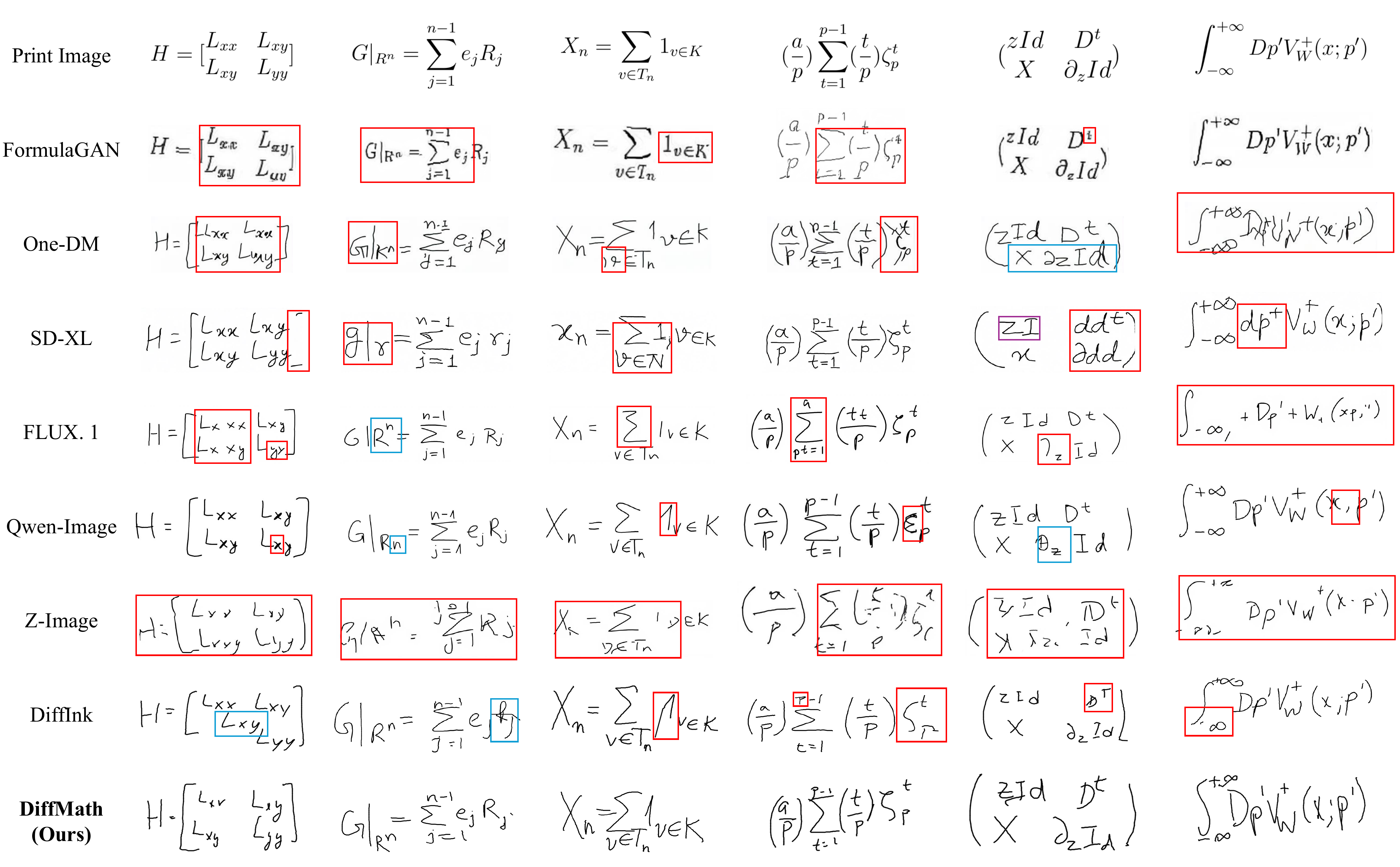}
        \caption{Comparison on \textbf{more complex} expressions.}
    \end{subfigure}
    
    \caption{
    \textbf{More qualitative comparisons with SOTA methods.} 
    Red boxes highlight atomic content errors (incorrect symbols or artifacts), blue boxes indicate structural misalignments, and purple boxes denote missing symbols. \textbf{DiffMath (Ours)} demonstrates strong performance in \textbf{both simple and complex LaTeX scenarios}, maintaining high fidelity in symbol content and 2D topology.
    }
    
    \label{fig:more_results}
\end{figure}

 As shown in Fig.\ref{fig:more_results}(a), DiffMath consistently generates handwritten mathematical expressions with more accurate symbol placement and stronger structural coherence than baseline generation methods. Even for formulas containing nested superscripts, fractions, roots, and matrix-like layouts, our method better preserves the intended mathematical structure while producing visually natural handwriting. Figure\ref{fig:more_results}(b) further showcases more challenging and diverse mathematical expressions generated by DiffMath, highlighting its ability to synthesize complex two-dimensional layouts with high structural fidelity and robustness.

\subsubsection{More Results Generated by DiffMath.}

\begin{figure}[t]
    \centering
    \includegraphics[width=1\linewidth]{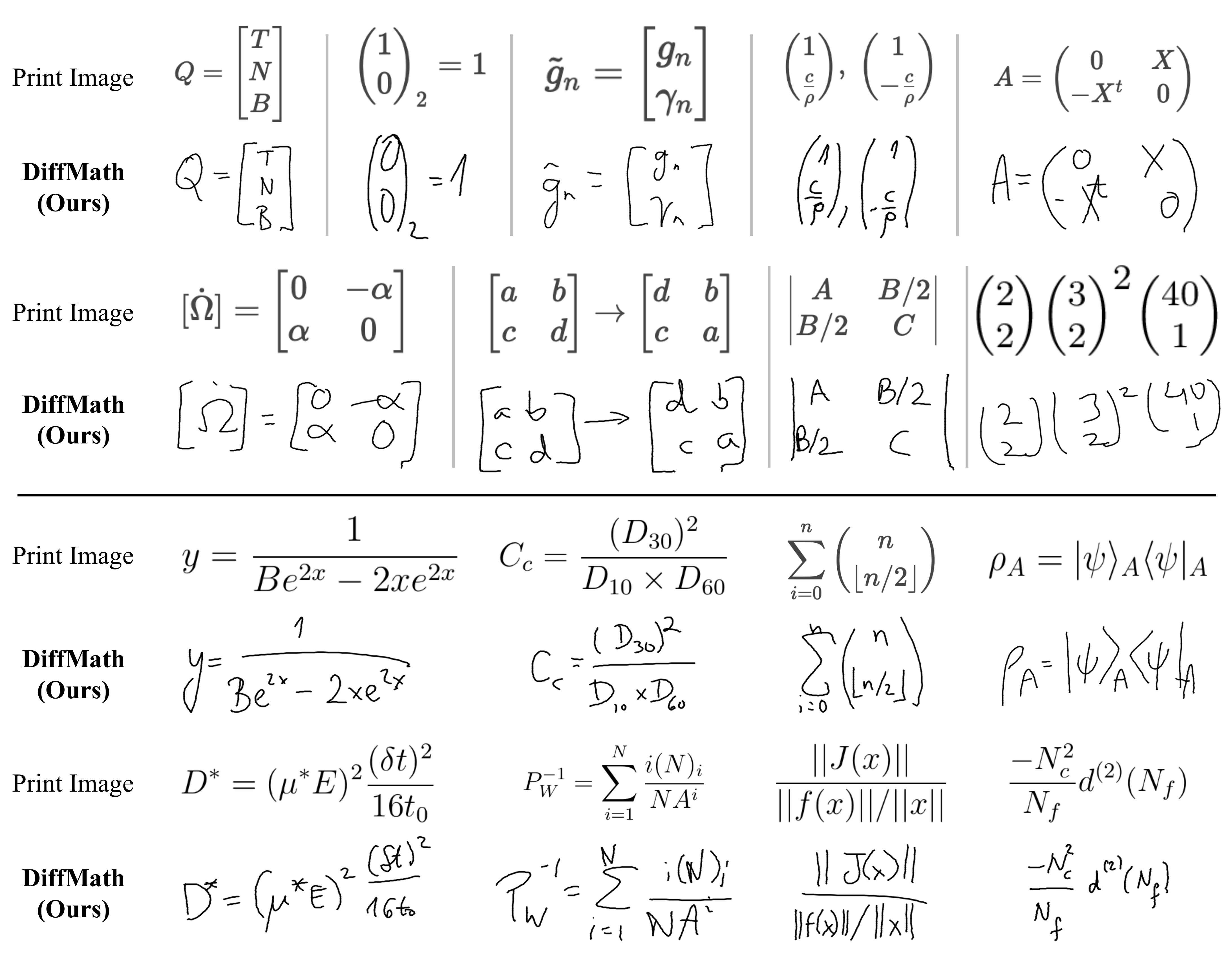}
    \caption{
    \textbf{Additional samples generated by DiffMath (Ours).}
    The upper part shows matrix expressions, while the lower part presents long formulas with dense symbols and complex local layouts. DiffMath consistently generates complex mathematical expressions with accurate symbols and correct structural relationships.
    }
    \label{fig:more_results_self}
\end{figure}

Furthermore, we present additional examples generated by DiffMath, as shown in Fig.~\ref{fig:more_results_self}. The upper part contains matrix expressions in various formats, while the lower part includes more complex expressions with numerous symbols and fine-grained details that are prone to errors. Even in these challenging cases, DiffMath maintains high accuracy in both symbol content and structural consistency.

\subsection{Failure Case Analysis}
\begin{figure}[t]
    \centering
    \includegraphics[width=1\linewidth]{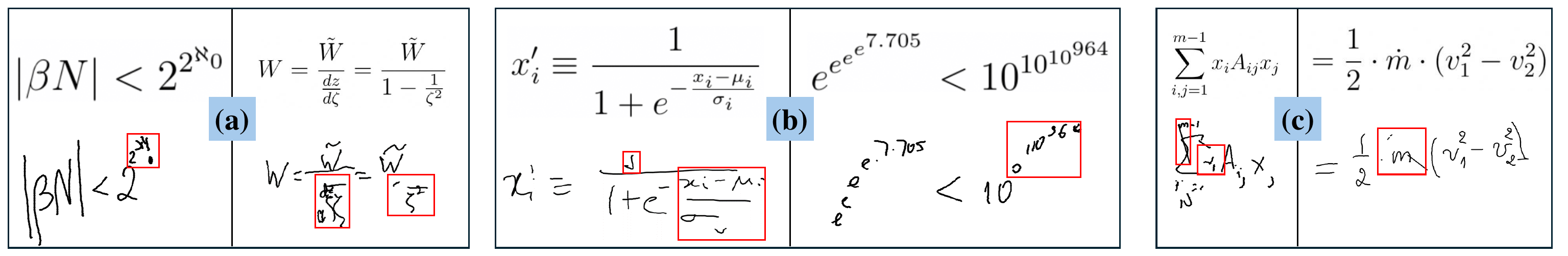}
     \caption{Failure cases of generated samples: (a) rare symbols, (b) complex nested structures, and (c) locally dense text regions.}
    \label{fig:failure_case}
\end{figure}

Fig.~\ref{fig:failure_case} presents representative failure cases. Most errors occur in LaTeX expressions with rare symbols, dense layouts, or deep nesting, where the model may omit small components, confuse similar symbols, or misplace superscripts, subscripts, and fraction elements. These cases indicate that compact and complex mathematical structures remain challenging for generation.

\end{document}